\documentclass[letterpaper, 10 pt, conference]{ieeeconf}  

\IEEEoverridecommandlockouts                              

\usepackage{graphics} 
\usepackage{epsfig} 
\usepackage{times} 
\usepackage{amsmath} 
\usepackage{amssymb}  
\usepackage{subfigure}
\usepackage{color}
\usepackage{multirow}
\usepackage{mathtools}
\usepackage{fancyvrb}
\usepackage{hyperref}

\title{\LARGE \bf
Accurate calibration of multi-perspective cameras from\\a generalization of the hand-eye constraint}

\author{Yifu Wang$^{*}$, Wenqing Jiang$^{*}$, Kun Huang, S\"oren Schwertfeger and Laurent Kneip
\thanks{$^{*}$ indicate equal contribution.}%
\thanks{Authors are from ShanghaiTech University; L. Kneip is also with the Shanghai Engineering Research Center of Intelligent Vision and Imaging. The authors would like to thank the funding sponsored by Natural Science Foundation of Shanghai (grant number: 22ZR1441300).}%
\thanks{Code: \url{github.com/MobilePerceptionLab/MultiCamCalib}}
}

\begin{document}

\maketitle
\thispagestyle{empty}
\pagestyle{empty}

\begin{abstract}
Multi-perspective cameras are quickly gaining importance in many applications such as smart vehicles and virtual or augmented reality. However, a large system size or absence of overlap in neighbouring fields-of-view often complicate their calibration. We present a novel solution which relies on the availability of an external motion capture system. Our core contribution consists of an extension to the hand-eye calibration problem which jointly solves multi-eye-to-base problems in closed form. We furthermore demonstrate its equivalence to the multi-eye-in-hand problem. The practical validity of our approach is supported by our experiments, indicating that the method is highly efficient and accurate, and outperforms existing closed-form alternatives.
\end{abstract}

\section{Introduction}

Robust and real-time 3D localization and exteroceptive perception have developed into core challenges to be solved towards the realization of many future robotics applications and intelligent mobile systems. The sensor that is commonly used in such contexts is a 360-degree Lidar. However, many applications may not be able to use Lidars for a number of reasons. For example, they are generally deemed too expensive in smart vehicle applications, in which we therefore often attempt to use a surround-view camera system---an option that is commonly available in modern vehicles for the purpose of generating panoramic views for parking assistance. Another example is given by virtual or augmented reality headsets, which have severe restrictions in terms of available energy budget and payload. Again, a common sensor alternative used in such applications is a multi-perspective camera. In summary, multi-perspective cameras are considered an interesting and affordable alternative for exteroceptive sensing in an increasing number of applications, such as \cite{furgale2013toward,wang2017scale, heng18, wang2020reliable, chen2020advanced, qin2020avp}.

Multi-perspective cameras offer the advantage of a potentially large fields-of-view, eventually reaching complete surround-view capabilities. This provides benefits in motion estimation accuracy and the ability to sense the environment in all directions around a platform. However, besides temporal synchronization and accurate intrinsics, the efficient and accurate use of multi-camera systems requires precise calibration of extrinsic camera parameters, thus enabling their treatment as generalized cameras \cite{pless03}. The present paper focuses on this problem.

The calibration of a multi-perspective camera system (MPC) is challenged by two factors. First, the system is often mounted on a large platform such as a passenger vehicle. This makes it difficult to apply the common camera calibration procedure of moving the cameras in front of a calibration target. Second---and more importantly in the context of this work---the various cameras of a multi-camera system often have very limited overlap in their fields of view, which makes it challenging to use classical methods relying on direct stereo vision constraints \cite{furgale2013unified}. A number of alternatives for the calibration of an MPC have therefore been presented in the literature, which are given by mirror-based calibration \cite{kumar2008simple, lebraly2010flexible, takahashi2012new, long2015simplified}, infra-structure based calibration \cite{ly2014extrinsic,choi2018automatic}, ego-motion or SLAM based calibration \cite{carrera2011slam,heng2013camodocal,heng2015leveraging,chen2019heterogeneous,ouyang2020online}, or hand-eye calibration \cite{Tsai,esquivel2007calibration,pachtrachai2018chess,zhi2017simultaneous}.

\begin{figure}[t]
	\centering
	\includegraphics[width=0.85\linewidth]{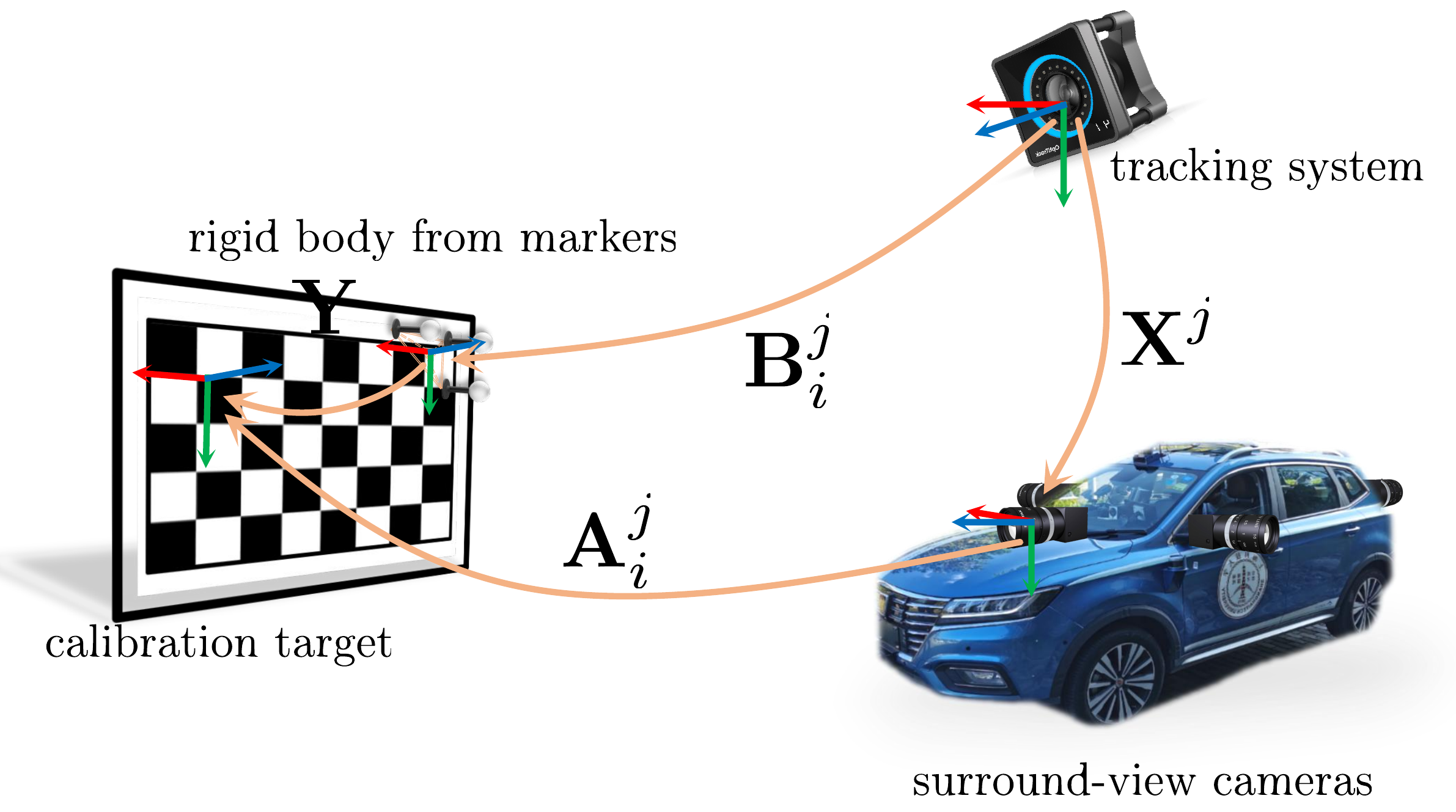}
	\caption{Proposed surround-view camera system calibration. The target is moved individually in the front of each camera, and its position is accurately measured by an external motion capture system.}
	\label{fig:concept}
	\vspace{-0.5cm}
\end{figure}

We present a highly accurate realization of a hand-eye calibration based method. As illustrated in Figure \ref{fig:concept}, our core idea consists of employing a motion capture system that is able to accurately measure the position of reflective markers attached to the calibration target. Each camera takes images of the target to find camera-to-target relative poses. Extrinsic parameters for each camera are then found by extracting their position inside the tracking system's reference frame. We solve this problem by applying transformation loop constraints, which only requires the additional solution of the target-to-marker-frame transformation. Note that the two unknown transformations in question are constant over time and may be recovered from multiple measurements of the target.
Our contributions are as follows:
\begin{itemize}
    \item We introduce a practical, simple and accurate extrinsic calibration procedure for non-overlapping multi-camera systems. Our method does not require the motion of cameras, which is especially suitable for cameras mounted on large devices. It furthermore does not need the cameras to be synchronized, which greatly reduces the complexity of the hardware setup.
	\item We present the first closed-form solution to multi-eye-to-base calibration, which can handle an arbitrary number of cameras. We furthermore demonstrate its equivalence to the multi-eye-in-hand problem.
	\item We successfully apply our framework to extrinsic camera calibration of multi-camera systems. Without using or requiring overlap, we prove that our algorithm achieves accuracy comparable to that of classical stereo calibration algorithms. It furthermore outperforms alternative closed-form hand-eye solvers which calibrate each camera individually.
\end{itemize}

\section{Related work}\label{sec:relatedWork}
Most approaches that aim at overcoming missing or reduced overlap between neighbouring fields-of-view of a multiple-camera system have been introduced in the introduction. While a number of successful works have already been presented, the conclusion is that most methods are either impractical or unable to achieve highly accurate results. Mirror-based calibration methods \cite{kumar2008simple, lebraly2010flexible, takahashi2012new, long2015simplified} require a single calibration target rendered visible in two cameras by using an additional mirror placed in front of one of the cameras. The requirement for a large perfectly planar mirror makes the approach impractical. Infra-structure based calibration methods \cite{ly2014extrinsic,choi2018automatic} depend on prior assumptions about the infrastructure (e.g. perfectly parallel lanes, vertical poles) being perfectly valid. Ego-motion or SLAM-based calibration approaches \cite{carrera2011slam,heng2013camodocal,heng2015leveraging,chen2019heterogeneous,ouyang2020online} recover the extrinsic parameters by aligning trajectories or performing extrinsic parameter-aware, large-scale bundle adjustment. The approaches easily suffer from common challenges in purely vision-based SLAM, which is drift, scale invariance, or a general lack of accuracy.

Hand-eye calibration \cite{Tsai,esquivel2007calibration,pachtrachai2018chess,zhi2017simultaneous} is a more practical method, which relies on geometric constraints between the poses of multiple rigidly-coupled cameras at different times. The method can be combined with traditional calibration tools in order to improve accuracy, such as observations of a known calibration target with known sizes. Hand-eye calibration is a classical geometric computer vision problem arising in the context of robot-mounted cameras, and can be formulated based on relative or absolute geometrical transformations constraints.  Relative transformation methods such as \cite{lebraly2011fast} generally appear in the form $\mathbf{AX} = \mathbf{XB}$, which is well analysed in \cite{shiu1987calibration}. Such methods make use of the relative transformations obtained from different cameras as well as the extrinsic camera-to-camera transformation parameters. Although such methods can achieve an accurate and efficient extrinsic calibration, they require all cameras to be fully synchronized, which increases the complexity of the hardware setup. Besides, such methods depend on controlled camera motion, which may be hard to execute if the cameras are mounted on a large platform. Another type of hand-eye calibration problem---denoted the hand-eye/robot-world calibration problem---relies on absolute geometric transformations. By constructing a loop constraint between the hand-eye and robot-arm coordinate frames, we can calculate the relative transformation between the fixed camera and the base coordinates. There are several approaches solving this problem for single camera cases, such as \cite{shah2013solving, li2010simultaneous}. Only few methods have extended the hand-eye/robot-world constraints towards multi-camera calibration. \cite{tabb2017solving} can only handle the case of multiple cameras moving in front of one and the same calibration target, while \cite{pedrosa2021general} presents a general approach to handle multi-agent (non-overlapping) hand-eye calibration. However, it is an optimization-based method that---owing to the non-linear, non-convex nature of the problem---highly depends on an accurate starting point.

\section{Foundations}\label{sec:foundations}
We start by introducing the notations and the geometry of our problem, summarize existing robot-world/hand-eye calibration techniques, and conclude with a motivation for our new, generalized solver.

\subsection{Notations and prior assumptions}
\begin{figure}[t!]
	\centering
	\subfigure[]{
		\label{fig:eye_on_hand}
		\includegraphics[width=0.485\linewidth]{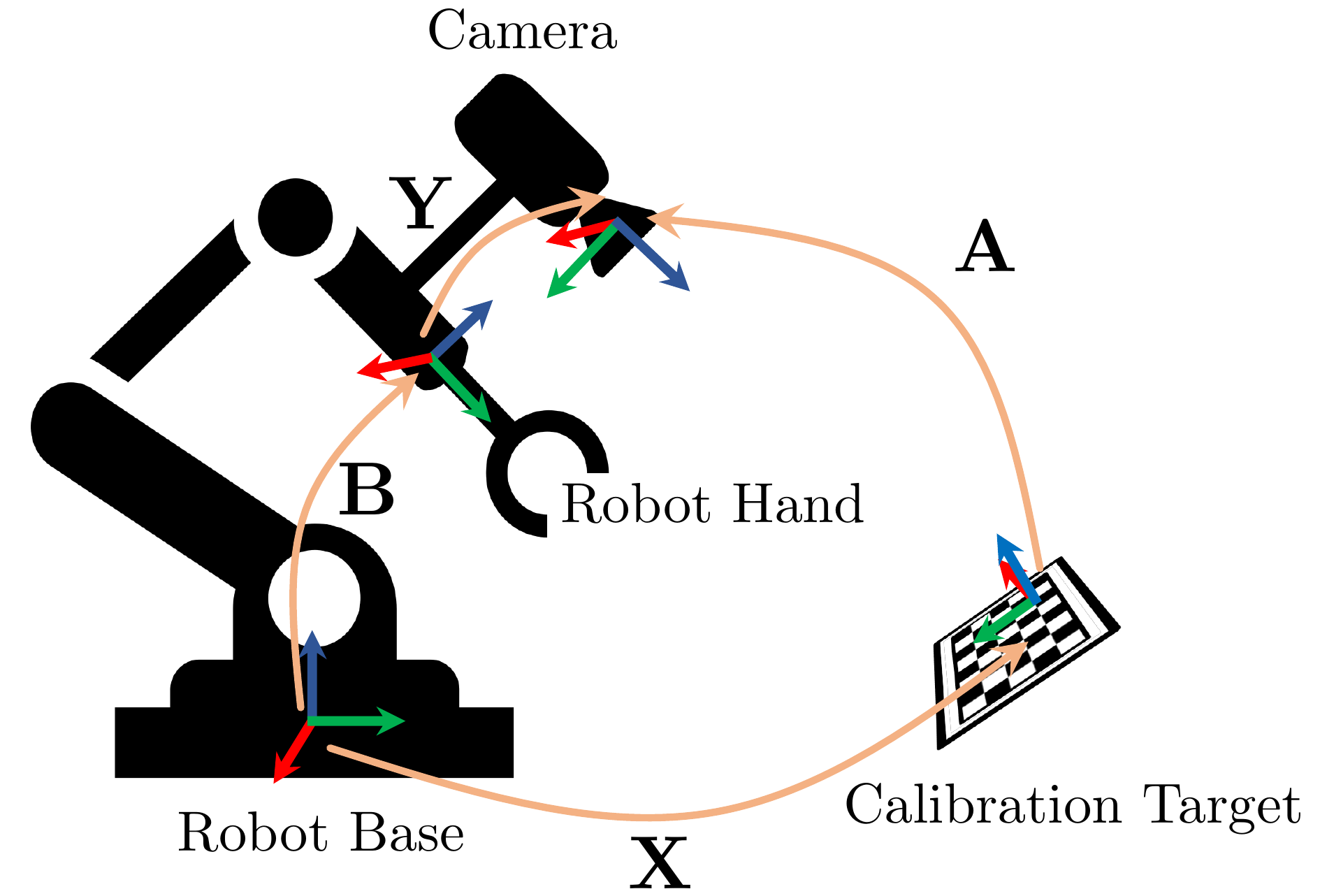}}
	\hspace{-0.3cm}
	\subfigure[]{
		\label{fig:eye_to_base}
		\includegraphics[width=0.485\linewidth]{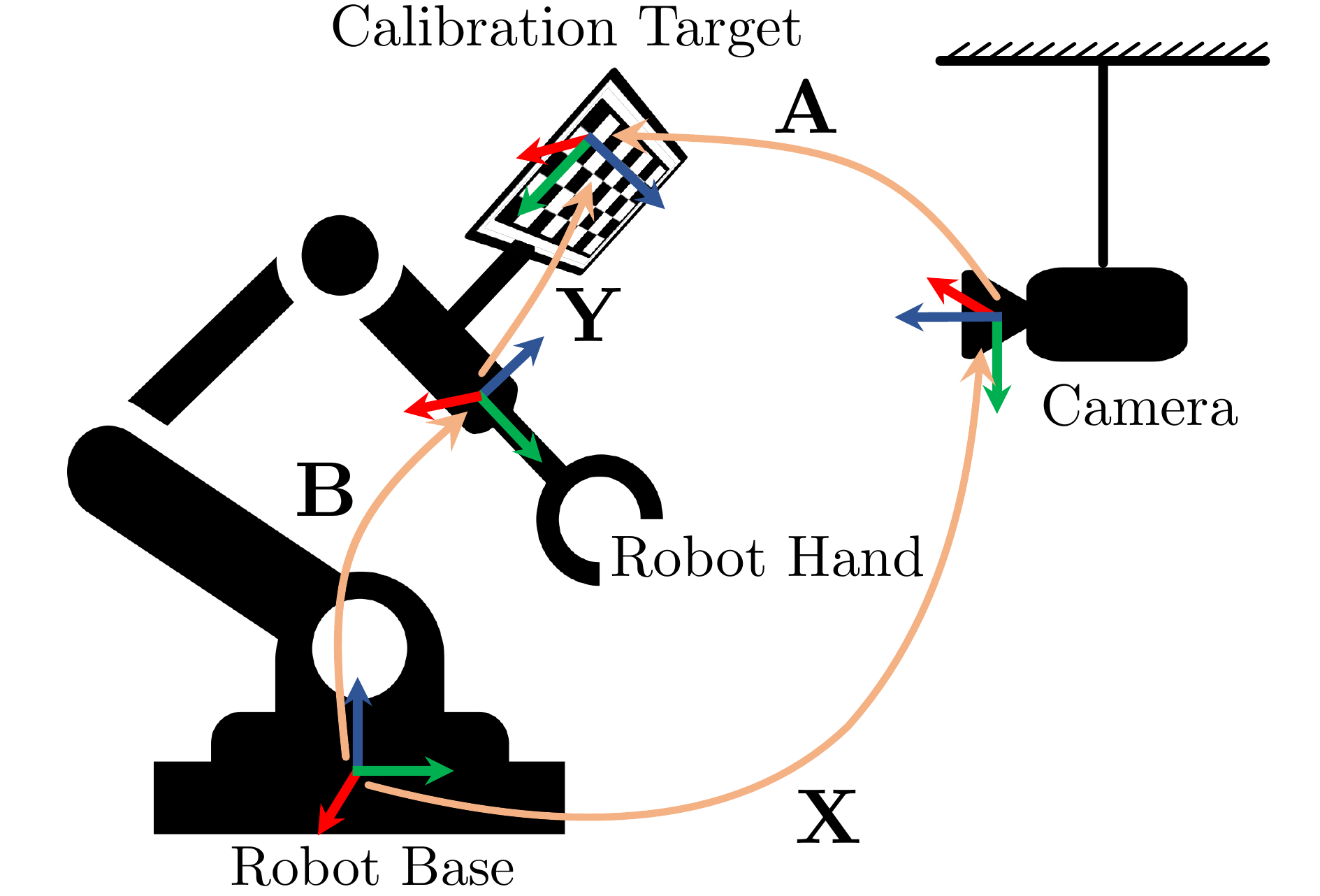}}
	\caption{Visualization of the hand-eye/robot-world calibration problem. Both (a) \textit{eye-on-hand} and (b) \textit{eye-to-base} cases are constrained by $\mathbf{A}\mathbf{X} = \mathbf{Y}\mathbf{B}$.}
	\vspace{-0.5cm}
\end{figure}
Hand–eye calibration problems can be divided into two cases. As shown in Figure \ref{fig:eye_on_hand} and \ref{fig:eye_to_base}, the \textit{eye-on-hand} case seeks the transformation between a rigidly attached end-effector (i.e. the hand) and camera (i.e. the eye), and the \textit{eye-to-base} case seeks the transformation between a fixed camera and the base of the robotic arm. The problems are equivalent from an algebraic perspective, and we use the \textit{eye-to-base} case for our further explanations. We assume that we have an intrinsically calibrated multi-camera system which is fixed in the world coordinate frame. Considering that we have a regular calibration target moving in front of each camera, we define $\mathbf{A}_i^j$
as the $i$th relative transformation of the fixed camera $j$ to the moving target, and let $\mathbf{B}_i^j$
%
be the corresponding transformation of the robot base coordinate frame to the hand coordinate frame, where $j\in\left\{1,\cdots,m\right\}$ and $i\in\left\{1,\cdots,N_j\right\}$. $\mathbf{A}_i^j$ can be easily solved by utilizing PnP methods \cite{collins2014infinitesimal,lepetit2009epnp,kneip2014upnp} with a known calibration pattern size, and $\mathbf{B}_i^j$ is directly obtained from the robot arm system---for which tracking system markers on the target are used lateron. Let $\mathbf{X}^j$
%
furthermore be the transformation from the robot-base coordinate frame to the fixed camera $j$. Finally, let $\mathbf{Y}$ be the transformation from the hand coordinate to the target coordinate frame. Note that $\mathbf{A}_i^j$, $\mathbf{B}_i^j$, $\mathbf{X}^j$ and $\mathbf{Y}$ are represented by a $3\times3$ rotation matrix $\mathbf{R}$ and a $3\times1$ translation vector $\mathbf{t}$.

\subsection{Brief review of hand-eye/robot-world calibration}
Note that the following exposition considers only a single camera, so index $j$ is dropped. The standard hand-eye/robot-world calibration constraint is given by
\begin{equation}
\mathbf{A}_i\mathbf{X} = \mathbf{Y}\mathbf{B}_i,
\label{eq:ax=zb}
\end{equation}
and most solvers solve the problem in two stages (first rotation, then translation):
\begin{gather}
\mathbf{R}_{\mathbf{A}_i}\mathbf{R_X} = \mathbf{R_Y}\mathbf{R}_{\mathbf{B}_i}
\label{eq:solve_RxRy}\\
\mathbf{R}_{\mathbf{A}_i}\mathbf{t_X} + \mathbf{t}_{\mathbf{A}_i}= \mathbf{R_Y}\mathbf{t}_{\mathbf{B}_i} + \mathbf{t_Y}.
\label{eq:solve_txty}
\end{gather}
The equations can be solved by either closed-form or iterative solutions. In this paper, we focus on closed-form solutions. As illustrated in \cite{shah2013solving}, we can apply the Kronecker product to represent (\ref{eq:solve_RxRy}) and (\ref{eq:solve_txty}) as linear equations, thus resulting in
\begin{gather}
\begin{pmatrix} -\mathbf{I} & \mathbf{R}_{\mathbf{B}_i}\otimes\mathbf{R}_{\mathbf{A}_i} \end{pmatrix} \begin{pmatrix} vec(\mathbf{R_Y}) \\ vec(\mathbf{R_X}) \end{pmatrix} = \mathbf{0}
\label{eq:solve_RxRy_linear}\\
\begin{pmatrix} \mathbf{I} & -\mathbf{R}_{\mathbf{A}_i} \end{pmatrix}
\begin{pmatrix} \mathbf{t_Y} \\ \mathbf{t_X} \end{pmatrix} = \mathbf{t}_{\mathbf{A}_i} - \mathbf{R_Y}\mathbf{t}_{\mathbf{B}_i},
\label{eq:solve_txty_linear}
\end{gather}
where $vec(\mathbf{R_Y})$ and $vec(\mathbf{R_X})$ are vectorized rotation matrices. Note that in practice, many such constraints are stacked into larger linear problems. Next, finding the nullspace for the first part of (\ref{eq:solve_RxRy_linear}) is equivalent to find an efficient and unique solution of
\begin{equation}\small
\begin{pmatrix} n\mathbf{I} & -\sum\limits_{i=1}^{n}\mathbf{R}_{\mathbf{B}_i}\otimes\mathbf{R}_{\mathbf{A}_i}\\
-\sum\limits_{i=1}^{n}\mathbf{R}^T_{\mathbf{B}_i}\otimes\mathbf{R}^T_{\mathbf{A}_i} & n\mathbf{I}\end{pmatrix} 
\begin{pmatrix} vec(\mathbf{R_Y}) \\ vec(\mathbf{R_X}) \end{pmatrix}
= \mathbf{0}.
\label{eq:origin_large}
\end{equation}
The normal equations of a linear system $\mathcal{A}\mathcal{X}=\mathcal{B}$ can be rephrased as $\mathcal{A}^T\mathcal{AX=A}^T\mathcal{B}$, which leads to another simplified solution for the second part of (\ref{eq:solve_txty_linear}) as well. The nullspace and thus the rotations $vec(\mathbf{R_Y})$ and $vec(\mathbf{R_X})$ are easily found by singular value decomposition of (\ref{eq:origin_large}). The translation is then recovered by substituting $\mathbf{R_Y}$ and $\mathbf{R_X}$ into the second constraint, and solved for example by applying Cholesky factorization. 

\section{Joint multi-agent hand-eye calibration}\label{sec:theory}

The standard solver is not well suited for a multi-camera system, as only a single camera can be calibrated at a time. Although we could recover the extrinsics between multiple cameras by using the individual \textit{eye-to-base} transformations, the solution would ignore the fact that the fixed hand-to-target transformation $\mathbf{R_Y}$ is shared by all calibration problems. It would be computed multiple times, each time suffering from errors due to a lack of constraints. The errors would furthermore propagate onto the parameters of interest, which are the \textit{eye-to-base} transformations. In the following, we therefore introduce a generalized hand-eye calibration solver that jointly solves for multiple extrinsic camera transformations as well as the hand-to-target transformation.
Our main contribution consists of extending the idea of \cite{shah2013solving} into a generalized solver. We start by seeing a new linear system, which enables the joint retrieval of a common hand-to-target rotation and multiple base-to-camera rotations. Next, we derive the translations again through a joint solution scheme. The section concludes with a proof of equivalence of the \textit{multi-eye-to-base} and the \textit{multi-eye-on-hand} cases, which both have interesting practical applications.

\subsection{Joint linear rotation estimation}
\label{sec:find_r}

Suppose that the $i$th pose of the $j$th camera with respect to the target ($\mathbf{A}_i^j$) and the corresponding transformation of the robot base frame to the hand frame ($\mathbf{B}_i^j$) have already been identified. The multi-camera hand-eye/robot-world calibration can be easily formulated as:
\begin{equation}
\mathbf{A}_i^j\mathbf{X}^j = \mathbf{Y}\mathbf{B}_i^j.
\label{eq:aixi=zbi}
\end{equation}
It is important to realize that once multiple such constraints are stacked into a large linear problem, the resulting equation is different from (\ref{eq:ax=zb}). (\ref{eq:aixi=zbi}) can be used to calculate $\mathbf{Y}$ as a unique variable for calibrating the complete, generalized camera system. Stacking all pose measurements, the rotation constraint becomes
\begin{equation}
\left\{
\begin{aligned}
\mathbf{R}_{\mathbf{A}_1^1}\mathbf{R}_{\mathbf{X}^1} &= \mathbf{R_Y}\mathbf{R}_{\mathbf{B}_1^1} \\
\mathbf{R}_{\mathbf{A}_2^1}\mathbf{R}_{\mathbf{X}^1} &= \mathbf{R_Y}\mathbf{R}_{\mathbf{B}_2^1} \\
&\cdots\\
\mathbf{R}_{\mathbf{A}_{N_m}^m}\mathbf{R}_{\mathbf{X}^m} &= \mathbf{R_Y}\mathbf{R}_{\mathbf{B}_{N_m}^m} \\
\end{aligned}
\right. .
\label{eq:solve_multiRxRy}
\end{equation}
Inspired by the definition of (\ref{eq:solve_RxRy_linear}), all sub-constrains in (\ref{eq:solve_multiRxRy}) can now be grouped into the linear problem
\begin{equation}
\begin{psmallmatrix*}\small
-\mathbf{I} & \mathbf{R}_{\mathbf{B}_1^1}\otimes\mathbf{R}_{\mathbf{A}_1^1} &        &  \\
-\mathbf{I} & \mathbf{R}_{\mathbf{B}_2^1}\otimes\mathbf{R}_{\mathbf{A}_2^1} &        &  \\
\\ & & \cdots &  \\ \\
-\mathbf{I} & & & \mathbf{R}_{\mathbf{B}_{N_m}^m}\otimes\mathbf{R}_{\mathbf{A}_{N_m}^m} \\
\end{psmallmatrix*}
\begin{psmallmatrix*}[c]
vec(\mathbf{R_Y}^{\;}) \\ vec(\mathbf{R}_{\mathbf{X}^1}) \\ \\ \cdots  \\ \\ vec(\mathbf{R}_{\mathbf{X}^m})
\end{psmallmatrix*}
=\mathbf{0}.
\label{eq:solve_multiRxRy_linear}
\end{equation}
In the spirit of (\ref{eq:origin_large}), we can again find an efficient and unique solution to the homogeneous linear system (\ref{eq:solve_multiRxRy_linear}) by moving to
\begin{equation}\small
\begin{pmatrix}
N_1\mathbf{I} & -\mathbf{K}_1 &  &  \\
\cdots &  & \cdots &  \\
N_m\mathbf{I} &  &  & -\mathbf{K}_m \\
-{\mathbf{L}_1} & N_1\mathbf{I} &  &  \\
\cdots &  & \cdots &  \\
-{\mathbf{L}_m} &  &  & N_m\mathbf{I} \\
\end{pmatrix} 
\begin{pmatrix*}[c]
vec(\mathbf{R_Y}^{\;}) \\ vec(\mathbf{R}_{\mathbf{X}^1}) \\ \cdots \\ vec(\mathbf{R}_{\mathbf{X}^m})
\end{pmatrix*}=\mathcal{U}\mathcal{X}=\mathbf{0},\nonumber
\label{eq:new_large}
\end{equation}
where
\begin{equation}
\mathbf{K}_j = \sum\limits_{i=1}^{N_j}\mathbf{R}_{\mathbf{B}^j_i}\otimes\mathbf{R}_{\mathbf{A}^j_i}\; ,
\mathbf{L}_j = \sum\limits_{i=1}^{N_j}\mathbf{R}^T_{\mathbf{B}^j_i}\otimes\mathbf{R}^T_{\mathbf{A}^j_i}.
\end{equation}
Note that $N_j$ is the number of pose measurements for the $j$th camera. The nullspace of $\mathcal{U}$ can still be efficiently computed by singular value decomposition. The exact solution
of $\mathcal{U}\mathcal{X}=0$ is given by the column of the right-hand nullspace matrix $\mathbf{V}$ corresponding to the smallest singular value. In order to recover the rotation matrices $\mathbf{R_Y}$ and $\mathbf{R}_{\mathbf{X}^j}$, we de-vectorize the solution and obtain the $3\times3$ matrices $\mathbf{M}_{\mathbf{X}^j} = vec^{-1}(\mathbf{R}_{\mathbf{X}^j})$ and $\mathbf{M}_{\mathbf{Y}} =vec^{-1}(\mathbf{R_Y})$. In order to ensure that $\mathbf{M}_{\mathbf{X}^j}$ and $\mathbf{M}_{\mathbf{Y}}$ both satisfy the side-constraints of $SO3$ elements, we conclude with a normalization. We first obtain
\begin{equation}
\left\{
\begin{aligned}
&\mathbf{R}_{\mathbf{X}^j} = \text{sign}(\mathbf{M}_{\mathbf{X}^j})\text{det}(\mathbf{M}_{\mathbf{X}^j})^{-\frac{1}{3}}\mathbf{M}_{\mathbf{X}^j}\\
&\mathbf{R_Y^{\quad}} = \text{sign}(\mathbf{M_Y})\text{det}(\mathbf{M_Y})^{-\frac{1}{3}}\mathbf{M_Y}
\end{aligned}
\right. ,
\end{equation}
which ensures that we have right-hand matrices of determinant 1. Finally we orthogonalize the matrices by using SVD.
\subsection{Recovery of translations}
\label{sec:find_t}
In order to recover the translation $\mathbf{t_Y}$ and $\mathbf{t}_{\mathbf{X}^j}$, we start by substituting the rotation matrices $\mathbf{R_Y}$ and $\mathbf{R}_{\mathbf{X}^j}$ from Sec. \ref{sec:find_r} into equation (\ref{eq:solve_txty}):
\begin{equation}
\left\{
\begin{aligned}
\mathbf{R}_{\mathbf{A}_1^1}\mathbf{t}_{\mathbf{X}^1} + \mathbf{t}_{\mathbf{A}_1^1} &= \mathbf{R_Y}\mathbf{t}_{\mathbf{B}_1^1} + \mathbf{t_Y}\\
\mathbf{R}_{\mathbf{A}_2^1}\mathbf{t}_{\mathbf{X}^1} + \mathbf{t}_{\mathbf{A}_2^1} &= \mathbf{R_Y}\mathbf{t}_{\mathbf{B}_2^1} + \mathbf{t_Y}\\
&\cdots\\
\mathbf{R}_{\mathbf{A}_{N_m}^m}\mathbf{t}_{\mathbf{X}^m} + \mathbf{t}_{\mathbf{A}_{N_m}^m} &= \mathbf{R_Y}\mathbf{t}_{\mathbf{B}_{N_m}^m} + \mathbf{t_Y}\\
\end{aligned}
\right. .
\label{eq:solve_multitxty}
\end{equation}
By grouping all sub-constrains in (\ref{eq:solve_multitxty}), we obtain the linear system:
\begin{equation}
\begin{psmallmatrix*}\small
\mathbf{I} & -\mathbf{R}_{\mathbf{A}_1^1} &        &  \\
\mathbf{I} & -\mathbf{R}_{\mathbf{A}_2^1} &        &  \\
           &                                         & \cdots &  \\
\mathbf{I} & &        &-\mathbf{R}_{\mathbf{A}_{N_m}^m}   \\
\end{psmallmatrix*}
\begin{psmallmatrix*}[c]
\mathbf{t_Y}^{\;} \\ \mathbf{t}_{\mathbf{X}^1} \\ \\ \cdots \\ \\ \mathbf{t}_{\mathbf{X}^m}
\end{psmallmatrix*}
=\begin{psmallmatrix*}
\mathbf{t}_{\mathbf{A}_1^1} - \mathbf{R_Y}\mathbf{t}_{\mathbf{B}_1^1} \\ \\ \cdots \\ \\ \mathbf{t}_{\mathbf{A}_{N_m}^m} - \mathbf{R_Y}\mathbf{t}_{\mathbf{B}_{N_m}^m}
\end{psmallmatrix*}.
\label{eq:solve_multitxty_linear}
\end{equation}
$\mathbf{t}_{\mathbf{X}^j}$ and $\mathbf{t}_{\mathbf{Y}}$ are the solutions to the non-homogeneous linear system $\mathcal{A}\mathcal{X}=\mathcal{B}$, which can again be brought into the normal form $\mathcal{A}^T\mathcal{A}\mathcal{X}=\mathcal{A}^T\mathcal{B}$ and solved by standard techniques such as Cholesky factorization. 

The final \textit{eye-to-base} transformation for each camera ${\mathbf{X}^j}^{-1}$ is given by ${\mathbf{X}^j}^{-1} = \begin{bmatrix*} {\mathbf{R}_{\mathbf{X}^j}}^T & -{\mathbf{R}_{\mathbf{X}^j}}^T\mathbf{t}_{\mathbf{X}^j} \\ 0 & 1\end{bmatrix*}$. Extrinsics between cameras in the system can be easily computed by using the chain of transformations
\begin{equation}
\mathbf{T}_{0,j} = {\mathbf{X}^{0}}^{-1}\mathbf{X}^{j},
\end{equation}
where $\mathbf{T}_{0,j}$ denotes the relative transformation from camera $j$ to camera $0$. We generally use the first camera as the reference frame with respect to which all other extrinsic camera poses are expressed.

\subsection{Extension towards the multi-eye-on-hand case}

The previous subsections have presented a novel closed-form solver for the \textit{multi-eye-to-base} constraint. A very similar problem is given by the 
\textit{multi-eye-on-hand} case, in which we have multiple cameras mounted on the robot arm, and they observe only a single target. In this scenario, we have multiple hand-to-eye transformations $\mathbf{Y}^j$, but only a single base-to-target transformation $\mathbf{X}$. The basic constraint of this problem is
\begin{equation}
    \mathbf{A}_i^j\mathbf{X} = \mathbf{Y}^j\mathbf{B}_i^j.
\end{equation}
It is interesting to see that this equation is of algebraically identical form, as we may simply take the inverse on either side
\begin{equation}
{\mathbf{B}_i^j}^{-1}{\mathbf{Y}^j}^{-1} = {\mathbf{X}}^{-1}{\mathbf{A}_i^j}^{-1},
\label{eq:aix=zibi}
\end{equation}
thus resulting in a form that is entirely similar to (\ref{eq:aixi=zbi}).
From a practical perspective, this means that our generalized solver can be easily used to handle two different cases. The first one is given by the \textit{multi-eye-to-base} case, which may be more relevant for calibrating larger scale systems that are hard to move. The second is given by the \textit{multi-eye-on-hand} calibration problem, which may be more relevant to calibrate smaller multi-camera setups. Note that, in the continuation, we will focus on calibrating a vehicle mounted surround-view camera system, hence the remainder of this paper will use the \textit{multi-eye-to-base} constraint.
\begin{figure}[t]
	\centering
	\includegraphics[width=0.75\linewidth]{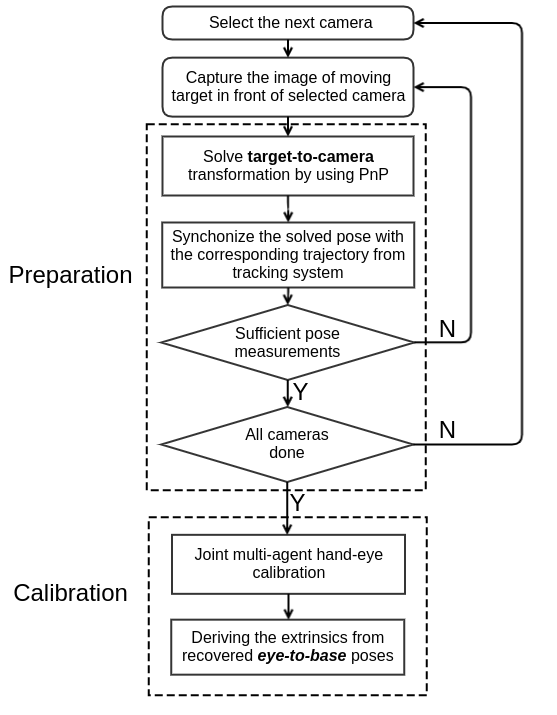}
	\caption{Overview of the proposed multi-agent hand-eye calibration pipeline for non-overlapping multi-camera systems.}
	\label{fig:flowchart}
	\vspace{-0.5cm}
\end{figure}
%
\section{Application to non-overlapping multi-camera systems}
\label{sec:application}
We evaluate our method on a non-overlapping multi-camera system, on which all cameras are facing into different directions and have no overlap in their fields-of-view. Many solutions to the case of regular hand-eye/robot-world calibration exist, including closed-form solvers \cite{li2010simultaneous} \cite{shah2013solving} and an iterative solution \cite{tabb2017solving}. The iterative solver can handle the \textit{eye-on-hand} case for multiple cameras, and solvers are otherwise restricted to the single camera case. To the best of our knowledge, we propose the first closed-form solution to hand-eye calibration for the \textit{eye-to-base} case which supports multiple non-overlapping cameras. The \textit{eye-to-base} case is particularly relevant in situations where all cameras are attached to a large-scale platform that cannot be easily manipulated. In our framework, the base-to-hand transformations $\mathbf{B}^j_i$ are given by a highly accurate external motion capture system. It keeps tracking the 3D position and orientation of a reflective marker frame (denoted \textbf{Rig}) mounted on the calibration target, which---in analogy to the hand-target transformation---requires the identification of an extrinsic transformation $\mathbf{Y}$ to the calibration target's reference frame. $\mathbf{A}^j_i$ and $\mathbf{X}^j$ are camera-to-target and tracking-system-to-camera transformations, respectively. A detailed overview of our framework is shown in the Figure \ref{fig:flowchart}.

We assume that the intrinsics of the cameras are pre-calibrated, thus we can directly use the PnP methods \cite{collins2014infinitesimal,lepetit2009epnp,kneip2014upnp}, to recover all $\mathbf{A}^j_i$'s. In order to obtain perfectly synchronized transformation measurements, each camera is hard synchronized with the tracking system. Note however that the calibration target can be moved in front of each camera individually, and thus the image sets for each camera can be recorded sequentially.

\section{Experimental Evaluation}
\label{sec:experiments}

In this section, we briefly introduce implementation details of our method and evaluate the performance on both synthetic and real data. Our solver depends on \textit{eye-to-base} hand-eye/robot-world calibration constraints and is designed for non-overlapping multi-camera systems. Our experiments therefore focus on a comparison against previous hand-eye/robot-world algorithms, which are closed-form hand-eye solutions \cite{shah2013solving} and \cite{li2010simultaneous}, and an iterative method \cite{tabb2017solving} designed for multi-camera systems. However, all methods above are not able to handle the \textit{multi-eye-to-base} case, thus we compute the \textit{eye-to-base} transformation for each camera individually. The extrinsics between cameras are each time derivied from a chain of transformations. We execute different comparative simulation experiments to evaluate accuracy and noise resilience, and evaluate the performance of the proposed method for a different number of cameras. We conclude with extrinsic calibration of both a surround-view camera system and an overlapping stereo camera with non-overlapping assumptions. Ground truth for the stereo camera calibration is delivered by a classical stereo calibration method. Implementations are made in C++ and use OpenCV \cite{opencv_library} for image processing and the solution of geometric problems. All experiments are conducted on an Intel Core i7 2.4 GHz CPU with 8GB RAM. 

\subsection{Error Metrics}
\label{sec:error_metrics}

Next we introduce the error metrics for comparing our solver against alternatives. Similar to \cite{tabb2017solving}, we use the hand-eye constraint (\ref{eq:ax=zb}) to represent rotation and translation errors in the absence of ground truth extrinsics:
\begin{equation}\small
\left\{
\begin{aligned}
e_{\mathbf{R}}&=\frac{1}{n}\frac{1}{m}\sum\limits_{j=0}^{m-1}\sum\limits_{i=0}^{n-1}angle((\mathbf{R}_{\mathbf{Y}}\mathbf{R}_{\mathbf{B}^j_i})^T(\mathbf{R}_{\mathbf{A}^j_i}\mathbf{R}_{\mathbf{X}^j}))\\
e_{\mathbf{t}}&=\frac{1}{n}\frac{1}{m}\sum\limits_{j=0}^{m-1}\sum\limits_{i=0}^{n-1}\|(\mathbf{R}_{\mathbf{A}^j_i}\mathbf{t}_{\mathbf{X}^j}+\mathbf{t}_{\mathbf{A}^j_i})-(\mathbf{R}_{\mathbf{Y}}\mathbf{t}_{\mathbf{B}^j_i}+\mathbf{t}_{\mathbf{Y}})\|
\end{aligned}
\right.
\end{equation}
Please note that we use averaged $\mathbf{R}_{\mathbf{Y}}$ and $\mathbf{t}_{\mathbf{Y}}$ computed from each camera when we are evaluating the solvers designed for the single camera case.
\subsection{Results on synthetic data}

In our simulation experiment, we generate a surround-view camera system that highly resembles the multi-camera system in our real experiments. It has four cameras pointing into all directions (cf. Figure \ref{fig:concept}). The cameras all lie in the same horizontal plane and have a distance between 0.4 and 0.65m away from the body origin. For the simulated dataset, we add up to 40 camera-to-target poses $\mathbf{B}_i^j$ for each camera, which are taken from a real sequence and thus have realistic values. We then generate $\mathbf{A}_i^j$ by using 
$\mathbf{A}_i^j = \mathbf{Y}\mathbf{B}_i^j{\mathbf{X}^j}^{-1}$.
Note that the iterative method in \cite{tabb2017solving} minimizes reprojection errors in their objective function and therefore requires a simulated camera model and point reprojections. Thus, we only compare our method against other state-of-the-art closed-form solvers in our simulation experiments. We analyze the performance of our method for different noise levels, and varying numbers of cameras and measurements per camera. In practical calibration procedures, the measurements are obtained from well-selected calibration images, thus we do not add any outliers in the simulation experiment. We finally report the accuracy based on above mentioned error metrics.
\begin{figure}[t]
	\centering
	\subfigure[Noise level]{
		\includegraphics[width=0.485\linewidth]{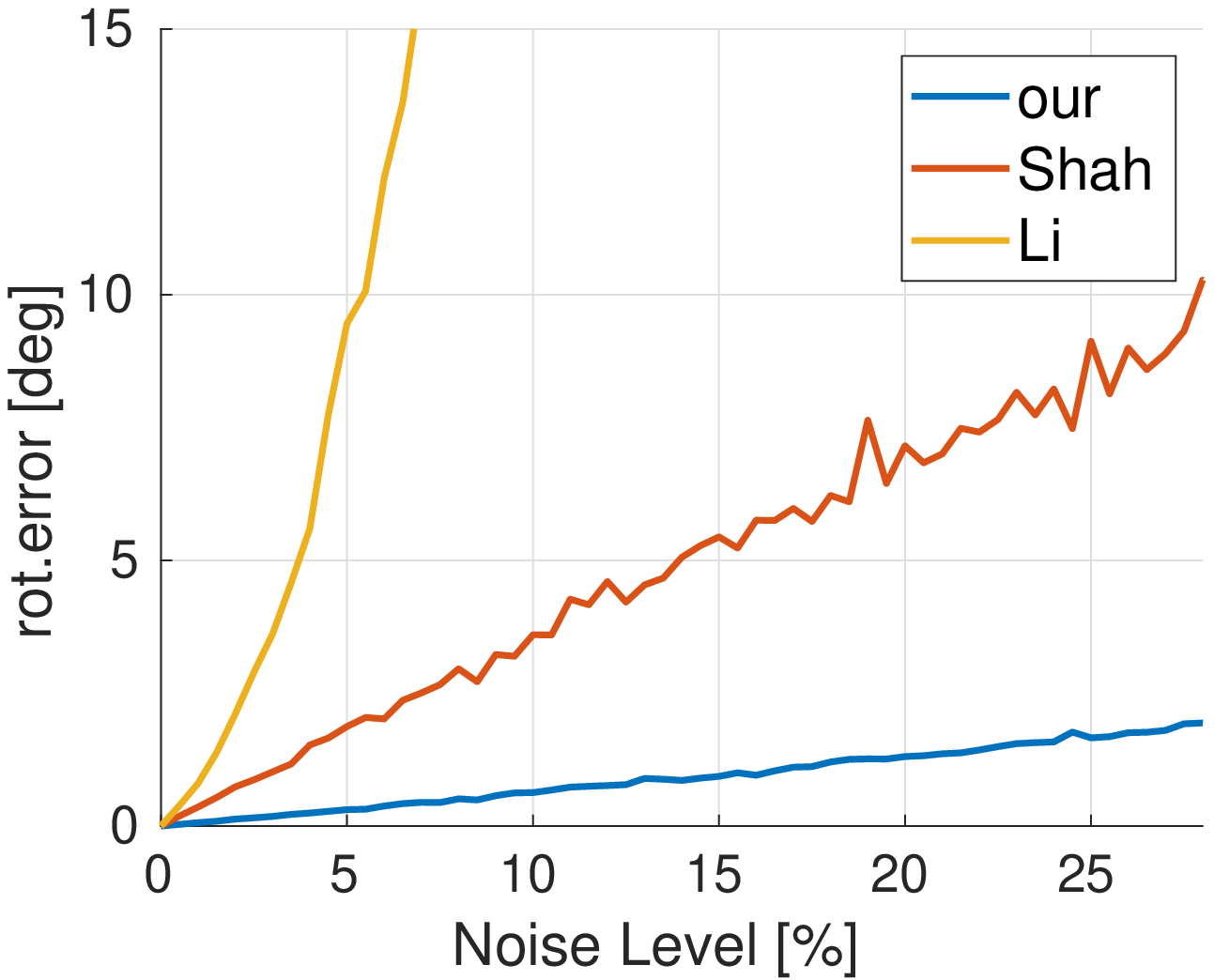}
		\includegraphics[width=0.485\linewidth]{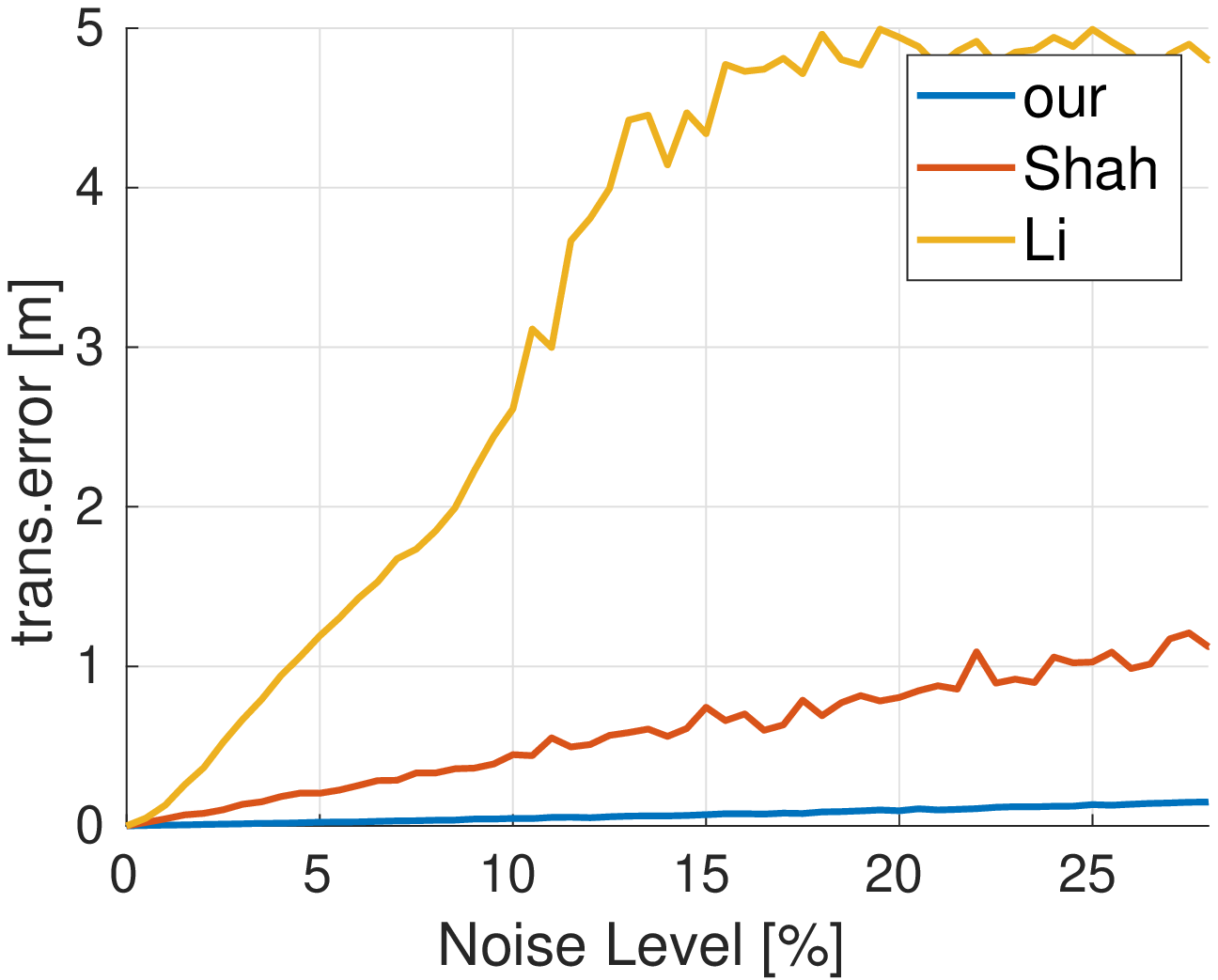}}
		\vspace{-0.1cm}
	\subfigure[Number of cameras]{
		\includegraphics[width=0.485\linewidth]{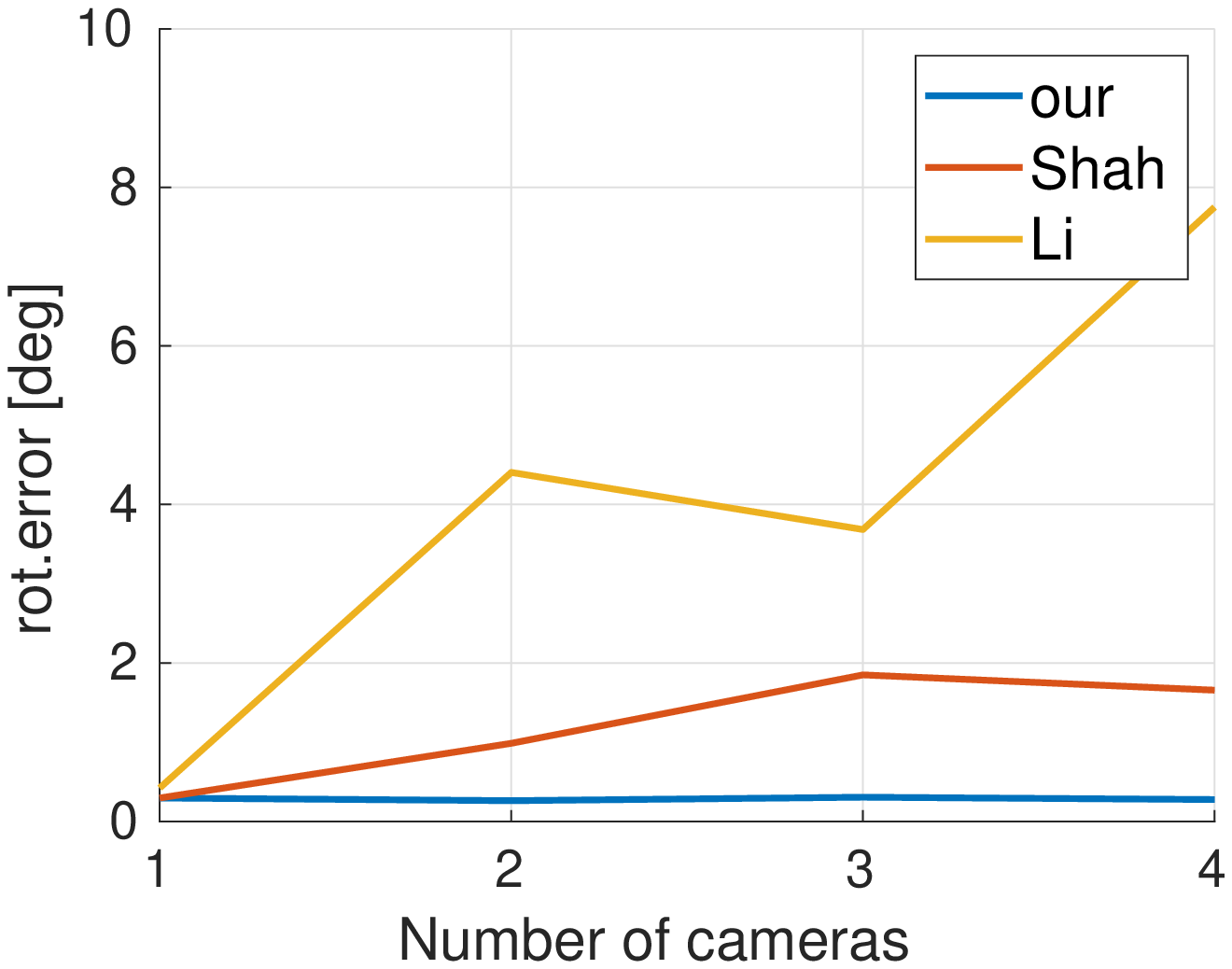}
		\includegraphics[width=0.485\linewidth]{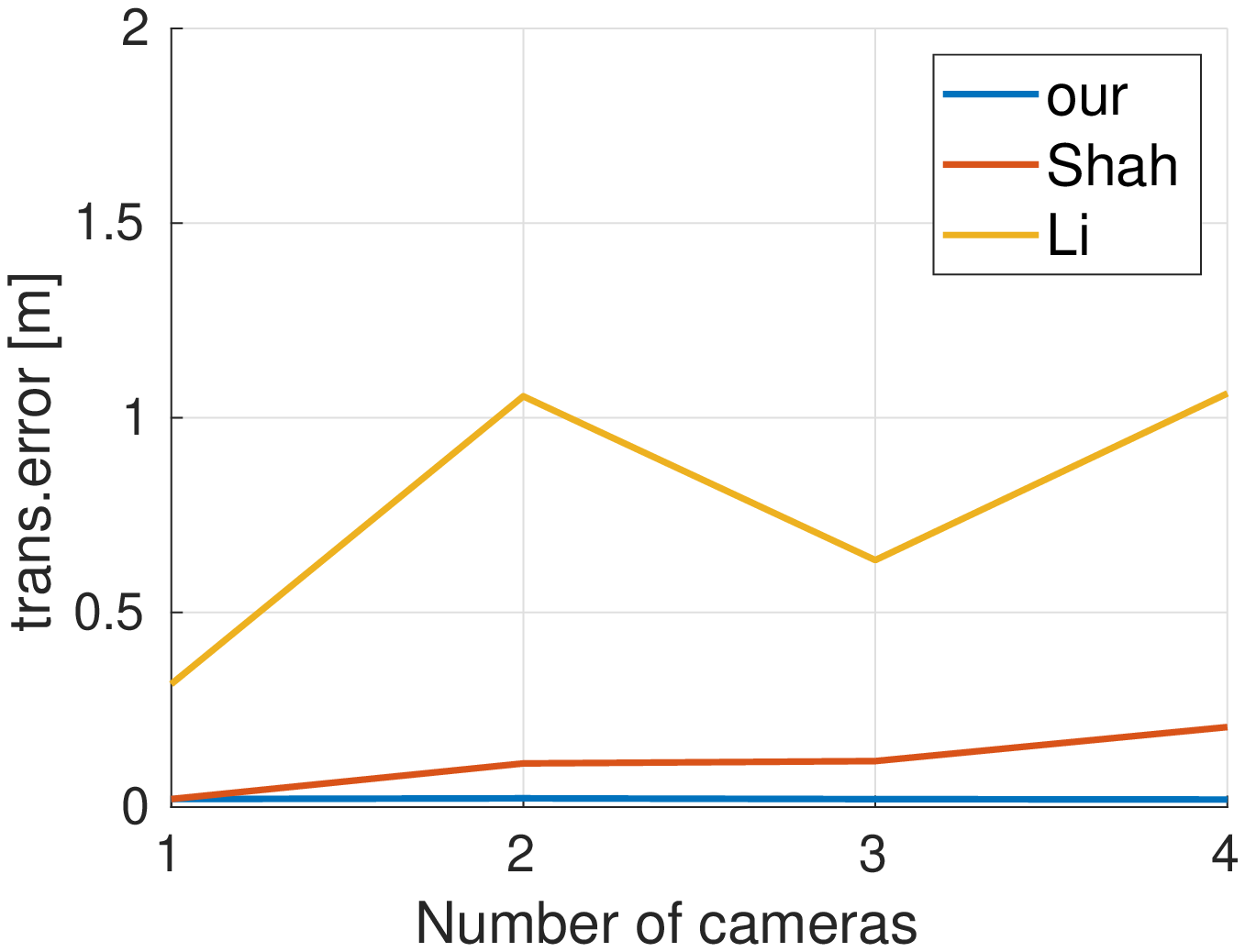} }
		\vspace{-0.1cm}
	\subfigure[Number of measurements]{
		\includegraphics[width=0.485\linewidth]{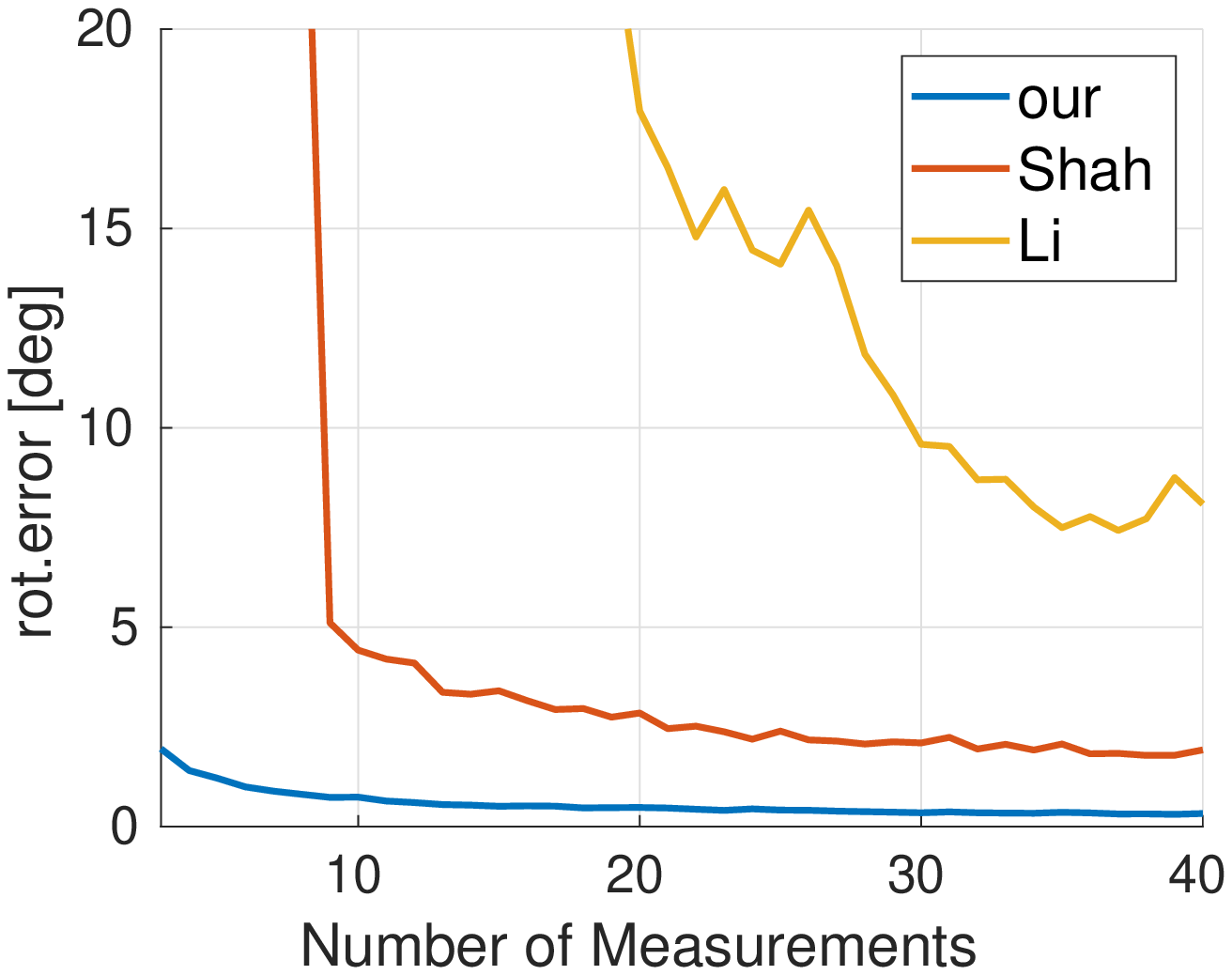}
		\includegraphics[width=0.485\linewidth]{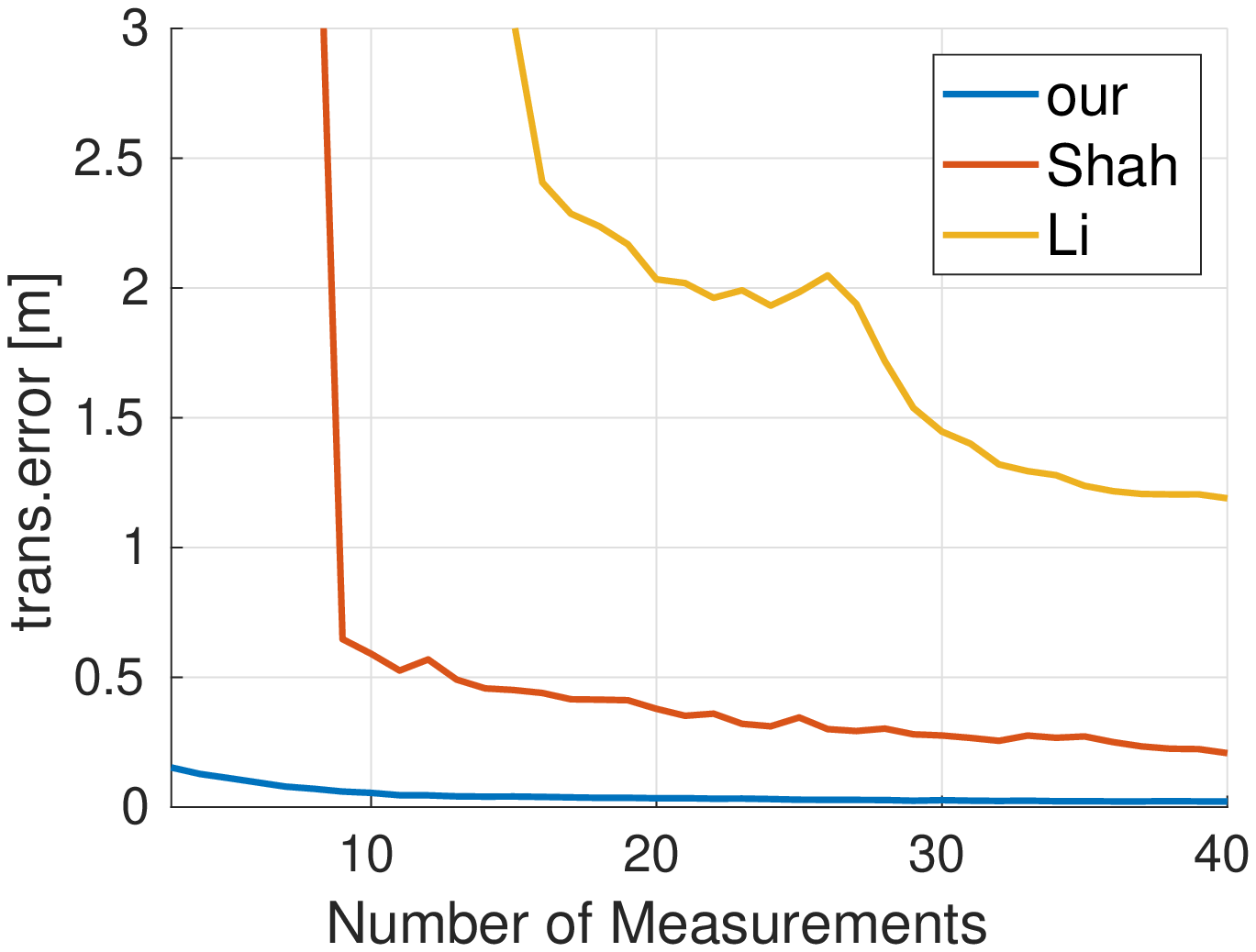} }
	\caption{Comparison between our proposed method \textbf{Our} and \textbf{Shah}'s and \textbf{Li}'s method for different perturbation factors. Each value is averaged over 100 random experiments. Details are provided in the text.}
	\label{fig:simulation}
	\vspace{-0.5cm}
\end{figure}
We compare our method against two alternative closed-form, hand-eye/robot-world methods, denoted (\textbf{Shah} \cite{shah2013solving} and \textbf{Li} \cite{li2010simultaneous}). We conduct three types of experiments:
\begin{itemize}
	\item \textit{Noise level}: 
	We use the full 4-camera system and 40 measurements for each camera. Noise is generated by taking a random faction of the absolute coordinates (up to 30\%), and adding it directly onto the measurements for both rotation and translation. Note that the camera system is placed in the center of the tracking system's reference frame, which is why the absolute poses---and thus also the random errors---have a relatively homogeneous distribution across the tracking system's area. As shown in Figure \ref{fig:simulation}(a), for both \textbf{Shah} and \textbf{Li}, the noise addition leads to a significant increase of the errors, especially for \textbf{Li}. Our proposed method performs best in terms of both rotational and translational errors.
	\item \textit{Number of cameras}: We fix the noise ratio to 5\% and vary the number of cameras to be calibrated from 1 to 4. The results are illustrated in Figure \ref{fig:simulation}(b). As expected, our method is well-suited for calibrating multi-camera systems. Adding more cameras will not decrease accuracy, while for \textbf{Shah} and \textbf{Li}, the errors increase with the number of cameras in the system.
	\item \textit{Number of measurements}:  We keep using the full 4-camera system and fix the noise ratio to 5\%. We vary the number of measurements for each camera from 3 to 40. Note that 3 is the minimum number of pose measurements for a single hand-eye/robot-world calibration solver. The results are indicated in Figure \ref{fig:simulation}(c). As can be observed, using more pose measurements leads to a large reduction of errors for all methods, and our method maintains a higher level of accuracy than the alternatives, which also proves that well-distributed pose measurements can significantly improve the calibration accuracy.
\end{itemize}
To conclude, we compare the computational efficiency of different methods. All methods process 160 measurements for 4 cameras in total. Our method's processing time is 5.54ms, \textbf{Shah} uses 3.58ms, \textbf{Li} uses 14.77ms. As mentioned in \cite{tabb2017solving}, the linear \textbf{Li} method will decompose an $8n\times16$ matrix using SVD, where $n$ is the number of measurements, thus it is 3 times slower than our method and \textbf{Shah}. Our method has comparable efficiency to the fastest alternative.

\subsection{Experiments on Real Data}
\label{sec:real_exp}

\begin{figure}[t]
	\vspace{0.2cm}
	\centering
	\subfigure[]{
		\label{fig:realSystem}
		\includegraphics[width=0.485\linewidth]{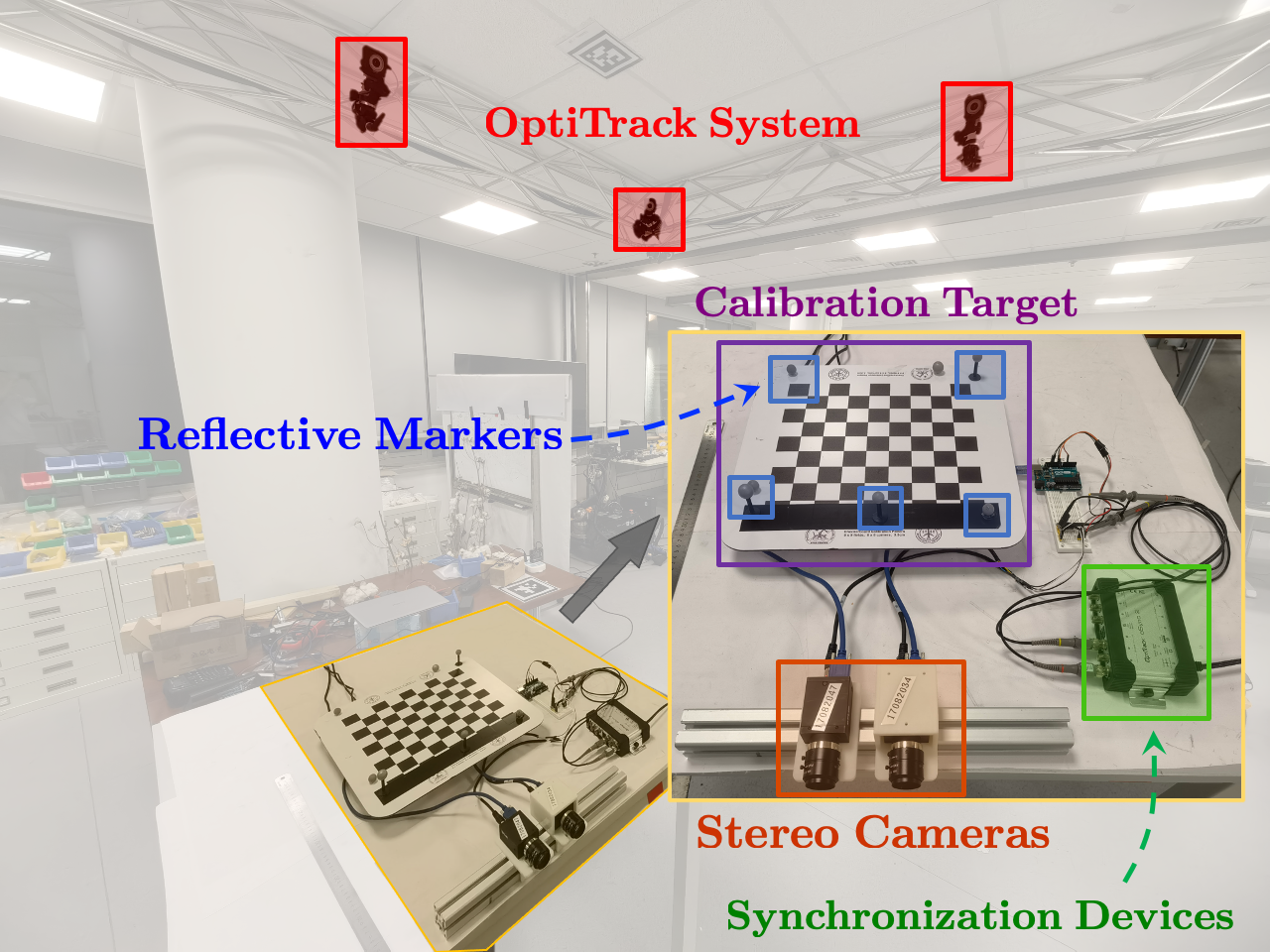}}
	\hspace{-0.3cm}
	\subfigure[]{
		\label{fig:real4cam}
		\includegraphics[width=0.485\linewidth]{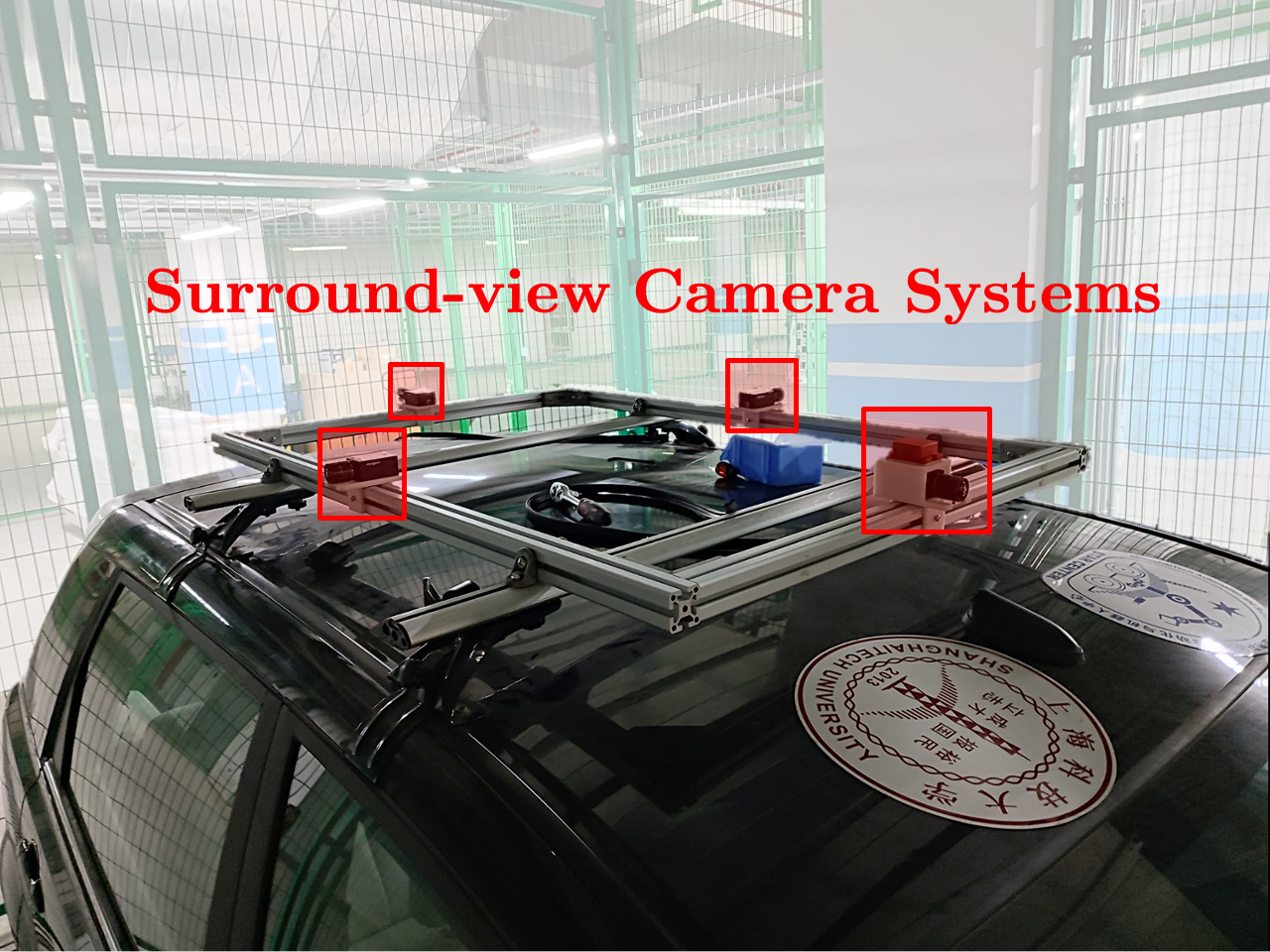}}
	\caption{Multi-camera systems as analyzed in this paper. (a) a stereo setup with overlapping fields-of-view and (b) a surround-view camera system.}
	\vspace{-0.4cm}
\end{figure}
In order to demonstrate the performance of our algorithm on real platforms, we apply it to two multi-camera systems given by a surround-view camera system and a stereo setup with overlapping fields-of-view (cf. illustrated in Figure \ref{fig:realSystem} and \ref{fig:real4cam}). Note that the surround-view camera system is mounted on a mobile rig and first calibrated inside the lab. Only after the calibration is finished, the entire frame is installed on top of the vehicle. The stereo setup allows us to compare our calibration results against a classical stereo calibration method , denoted (\textbf{GT} \cite{furgale2013unified}). 

Table \ref{tab:exps} shows our results on both surround-view and stereo camera systems and compares them against all alternatives. The retrieved extrinsic parameters from all methods are listed in Table \ref{tab:compare_to_GT}. The following is worth noting:
\begin{itemize}
	\item We carefully select 80 calibration images for the surround-view camera setup and 50 for the stereo setup. We add the iterative methods introduced by \cite{tabb2017solving} as alternatives and select the two best methods based on geometric constraints and reprojection errors respectively, which are \textbf{c1-Euler-separable} and \textbf{c2-Euler-separable} in \cite{tabb2017solving}. As shown in Table \ref{tab:exps}, all algorithms successfully complete the calibration without gross errors in rotations, while the translational errors are much more obvious due to error propagation. This is especially true for \textbf{Li} and \textbf{Tabb} methods. Similar to our simulation experiments, our method again outperforms all alternatives including iterative methods.
	\item A direct comparison of the retrieved extrinsic parameters is shown in the Table \ref{tab:compare_to_GT}. As can be observed, our calibration result with non-overlapping assumption is nearly as accurate as classical algorithms exploiting overlap in the fields-of-view.
\end{itemize}
\begin{table}[hb]
	\centering
	\caption{Comparison of methods using the error metrics described in Section \ref{sec:error_metrics} on surround-view and stereo camera systems.}
	\begin{tabular}{ccccc}
		\hline
		& \multicolumn{2}{c}{surround-view dataset} & \multicolumn{2}{c}{stereo dataset} \\ \cline{2-5} 
		Method    & $e_{\mathbf{R}}$ (deg)& $e_{\mathbf{t}}$ (m)& $e_{\mathbf{R}}$ (deg)& $e_{\mathbf{t}}$ (m)         \\ \hline
		Shah \cite{shah2013solving}      & 2.184                & 0.072              & 3.229             & 0.175          \\
		Li \cite{li2010simultaneous}       & 3.070                & 0.534              & 6.773             & 1.880          \\
		Tabb \cite{tabb2017solving} (c1) & 2.138                & 0.124              & 3.687             & 0.791          \\
		Tabb \cite{tabb2017solving} (c2) & 2.137                & 0.106              & 3.551             & 0.714          \\ \hline
		Our       & \textbf{1.423}      & \textbf{0.035}      & \textbf{1.927}   & \textbf{0.086}  \\ \hline
	\end{tabular}
	\label{tab:exps}
	\vspace{-0.4cm}
\end{table}
\begin{table}[ht]
	\centering
	\caption{Comparison of extrinsic parameters against GT. [$\mathbf{R}$: {\upshape deg}, $\mathbf{t}$: {\upshape m}]}
	\begin{tabular}{ccccccc}
		\hline
		Method       & $\mathbf{t}_x$              & $\mathbf{t}_y$              & $\mathbf{t}_z$              & $yaw$ & $pitch$         & $roll$           \\ \hline
		Shah \cite{shah2013solving}         & -0.057         & -0.264         & 0.028          & 2.80            & 2.09          & 2.08           \\
		Li \cite{li2010simultaneous}           & 0.496          & 0.376          & -0.351         & 0.11                     & 7.43          & 9.13           \\
		Tabb \cite{tabb2017solving} (c1) & 0.228          & -0.166         & -0.019         & 3.24                     & 3.36          & 2.04           \\
		Tabb \cite{tabb2017solving} (c2) & 0.190          & -0.146         & -0.022         & 3.24                     & 3.35          & 2.05           \\
		Our & 0.078 & 0.005 & 0.006 & -0.45           & 0.19 & -1.59 \\ \hline
		\textbf{GT \cite{furgale2013unified}}           & \textbf{0.091}          & \textbf{0.000}          & \textbf{0.005}          & \textbf{1.51}                     & \textbf{-0.14}         & \textbf{0.52}           \\ \hline
	\end{tabular}
	\label{tab:compare_to_GT}
\end{table}
\section{Conclusion}

We present a novel calibration technique for non-overlapping multi-camera systems that relies on an external motion capture system. Our work stands in contrast with many prior iterative optimization schemes presented in the literature, as it is a closed-form solution which does not suffer from an inadequate starting point and can be solved very efficiently. High accuracy is achieved by our solution to the multi-eye-to-base problem, and we demonstrate its equivalence to the multi-eye-in-hand problem, both of which are extensions of the well-known hand-eye calibration constraint. The method is easy to re-implement, and thus of high practical value to the community.

{\small
\bibliographystyle{IEEEtran}
\bibliography{root}

\begin{thebibliography}{10}
\providecommand{\url}[1]{#1}
\csname url@rmstyle\endcsname
\providecommand{\newblock}{\relax}
\providecommand{\bibinfo}[2]{#2}
\providecommand\BIBentrySTDinterwordspacing{\spaceskip=0pt\relax}
\providecommand\BIBentryALTinterwordstretchfactor{4}
\providecommand\BIBentryALTinterwordspacing{\spaceskip=\fontdimen2\font plus
\BIBentryALTinterwordstretchfactor\fontdimen3\font minus
  \fontdimen4\font\relax}
\providecommand\BIBforeignlanguage[2]{{%
\expandafter\ifx\csname l@#1\endcsname\relax
\typeout{** WARNING: IEEEtran.bst: No hyphenation pattern has been}%
\typeout{** loaded for the language `#1'. Using the pattern for}%
\typeout{** the default language instead.}%
\else
\language=\csname l@#1\endcsname
\fi
#2}}

\bibitem{furgale2013toward}
P.~Furgale, U.~Schwesinger, M.~Rufli, W.~Derendarz, H.~Grimmett,
  P.~M{\"u}hlfellner, S.~Wonneberger, J.~Timpner, S.~Rottmann, B.~Li,
  \emph{et~al.}, ``Toward automated driving in cities using close-to-market
  sensors: An overview of the v-charge project,'' in \emph{2013 IEEE
  Intelligent Vehicles Symposium (IV)}.\hskip 1em plus 0.5em minus 0.4em\relax
  IEEE, 2013, pp. 809--816.

\bibitem{wang2017scale}
Y.~Wang and L.~Kneip, ``On scale initialization in non-overlapping
  multi-perspective visual odometry,'' in \emph{International Conference on
  Computer Vision Systems}.\hskip 1em plus 0.5em minus 0.4em\relax Springer,
  2017, pp. 144--157.

\bibitem{heng18}
L.~Heng, B.~Choi, Z.~Cui, M.~Geppert, S.~Hu, B.~Kuan, P.~Liu, R.~Nguyen, Y.~C.
  Yeo, A.~Geiger, G.~H. Lee, M.~Pollefeys, and T.~Sattler, ``Project
  {AutoVision}: Localization and 3{D} scene perception for an autonomous
  vehicle with a multi-camera system,'' \emph{arXiv}, vol. 1809.05477, 2018.

\bibitem{wang2020reliable}
Y.~Wang, K.~Huang, X.~Peng, H.~Li, and L.~Kneip, ``Reliable frame-to-frame
  motion estimation for vehicle-mounted surround-view camera systems,'' in
  \emph{2020 IEEE International conference on robotics and automation
  (ICRA)}.\hskip 1em plus 0.5em minus 0.4em\relax IEEE, 2020, pp. 1660--1666.

\bibitem{chen2020advanced}
H.~Chen, Z.~Yang, X.~Zhao, G.~Weng, H.~Wan, J.~Luo, X.~Ye, Z.~Zhao, Z.~He,
  Y.~Shen, \emph{et~al.}, ``Advanced mapping robot and high-resolution
  dataset,'' \emph{Robotics and Autonomous Systems}, vol. 131, p. 103559, 2020.

\bibitem{qin2020avp}
T.~Qin, T.~Chen, Y.~Chen, and Q.~Su, ``Avp-slam: Semantic visual mapping and
  localization for autonomous vehicles in the parking lot,'' in \emph{2020
  IEEE/RSJ International Conference on Intelligent Robots and Systems
  (IROS)}.\hskip 1em plus 0.5em minus 0.4em\relax IEEE, 2020, pp. 5939--5945.

\bibitem{pless03}
R.~Pless, ``Using many cameras as one,'' in \emph{Proceedings of the {IEEE}
  Conference on Computer Vision and Pattern Recognition ({CVPR})}, Madison, WI,
  USA, 2003, pp. 587--593.

\bibitem{furgale2013unified}
P.~Furgale, J.~Rehder, and R.~Siegwart, ``Unified temporal and spatial
  calibration for multi-sensor systems,'' in \emph{2013 IEEE/RSJ International
  Conference on Intelligent Robots and Systems}.\hskip 1em plus 0.5em minus
  0.4em\relax IEEE, 2013, pp. 1280--1286.

\bibitem{kumar2008simple}
R.~K. Kumar, A.~Ilie, J.-M. Frahm, and M.~Pollefeys, ``Simple calibration of
  non-overlapping cameras with a mirror,'' in \emph{2008 IEEE Conference on
  Computer Vision and Pattern Recognition}.\hskip 1em plus 0.5em minus
  0.4em\relax IEEE, 2008, pp. 1--7.

\bibitem{lebraly2010flexible}
P.~L{\'e}braly, C.~Deymier, O.~Ait-Aider, E.~Royer, and M.~Dhome, ``Flexible
  extrinsic calibration of non-overlapping cameras using a planar mirror:
  Application to vision-based robotics,'' in \emph{2010 IEEE/RSJ International
  Conference on Intelligent Robots and Systems}.\hskip 1em plus 0.5em minus
  0.4em\relax IEEE, 2010, pp. 5640--5647.

\bibitem{takahashi2012new}
K.~Takahashi, S.~Nobuhara, and T.~Matsuyama, ``A new mirror-based extrinsic
  camera calibration using an orthogonality constraint,'' in \emph{2012 IEEE
  Conference on Computer Vision and Pattern Recognition}.\hskip 1em plus 0.5em
  minus 0.4em\relax IEEE, 2012, pp. 1051--1058.

\bibitem{long2015simplified}
G.~Long, L.~Kneip, X.~Li, X.~Zhang, and Q.~Yu, ``Simplified mirror-based camera
  pose computation via rotation averaging,'' in \emph{Proceedings of the IEEE
  Conference on Computer Vision and Pattern Recognition}, 2015, pp. 1247--1255.

\bibitem{ly2014extrinsic}
D.~S. Ly, C.~Demonceaux, P.~Vasseur, and C.~P{\'e}gard, ``Extrinsic calibration
  of heterogeneous cameras by line images,'' \emph{Machine vision and
  applications}, vol.~25, no.~6, pp. 1601--1614, 2014.

\bibitem{choi2018automatic}
K.~Choi, H.~G. Jung, and J.~K. Suhr, ``Automatic calibration of an around view
  monitor system exploiting lane markings,'' \emph{Sensors}, vol.~18, no.~9, p.
  2956, 2018.

\bibitem{carrera2011slam}
G.~Carrera, A.~Angeli, and A.~J. Davison, ``Slam-based automatic extrinsic
  calibration of a multi-camera rig,'' in \emph{2011 IEEE International
  Conference on Robotics and Automation}.\hskip 1em plus 0.5em minus
  0.4em\relax IEEE, 2011, pp. 2652--2659.

\bibitem{heng2013camodocal}
L.~Heng, B.~Li, and M.~Pollefeys, ``Camodocal: Automatic intrinsic and
  extrinsic calibration of a rig with multiple generic cameras and odometry,''
  in \emph{2013 IEEE/RSJ International Conference on Intelligent Robots and
  Systems}.\hskip 1em plus 0.5em minus 0.4em\relax IEEE, 2013, pp. 1793--1800.

\bibitem{heng2015leveraging}
L.~Heng, P.~Furgale, and M.~Pollefeys, ``Leveraging image-based localization
  for infrastructure-based calibration of a multi-camera rig,'' \emph{Journal
  of Field Robotics}, vol.~32, no.~5, pp. 775--802, 2015.

\bibitem{chen2019heterogeneous}
H.~Chen and S.~Schwertfeger, ``Heterogeneous multi-sensor calibration based on
  graph optimization,'' in \emph{2019 IEEE International Conference on
  Real-time Computing and Robotics (RCAR)}.\hskip 1em plus 0.5em minus
  0.4em\relax IEEE, 2019, pp. 158--163.

\bibitem{ouyang2020online}
Z.~Ouyang, L.~Hu, Y.~Lu, Z.~Wang, X.~Peng, and L.~Kneip, ``Online calibration
  of exterior orientations of a vehicle-mounted surround-view camera system,''
  in \emph{2020 IEEE International Conference on Robotics and Automation
  (ICRA)}.\hskip 1em plus 0.5em minus 0.4em\relax IEEE, 2020, pp. 4990--4996.

\bibitem{Tsai}
R.~Tsai and R.~Lenz, ``A new technique for fully autonomous and efficient 3d
  robotics hand/eye calibration,'' \emph{IEEE Transactions on Robotics and
  Automation}, vol.~5, no.~3, pp. 345--358, 1989.

\bibitem{esquivel2007calibration}
S.~Esquivel, F.~Woelk, and R.~Koch, ``Calibration of a multi-camera rig from
  non-overlapping views,'' in \emph{Joint Pattern Recognition Symposium}.\hskip
  1em plus 0.5em minus 0.4em\relax Springer, 2007, pp. 82--91.

\bibitem{pachtrachai2018chess}
K.~Pachtrachai, F.~Vasconcelos, G.~Dwyer, V.~Pawar, S.~Hailes, and D.~Stoyanov,
  ``Chess—calibrating the hand-eye matrix with screw constraints and
  synchronization,'' \emph{IEEE Robotics and Automation Letters}, vol.~3,
  no.~3, pp. 2000--2007, 2018.

\bibitem{zhi2017simultaneous}
X.~Zhi and S.~Schwertfeger, ``Simultaneous hand-eye calibration and
  reconstruction,'' in \emph{2017 IEEE/RSJ International Conference on
  Intelligent Robots and Systems (IROS)}.\hskip 1em plus 0.5em minus
  0.4em\relax IEEE, 2017, pp. 1470--1477.

\bibitem{lebraly2011fast}
P.~L{\'e}braly, E.~Royer, O.~Ait-Aider, C.~Deymier, and M.~Dhome, ``Fast
  calibration of embedded non-overlapping cameras,'' in \emph{2011 IEEE
  international conference on robotics and automation}.\hskip 1em plus 0.5em
  minus 0.4em\relax IEEE, 2011, pp. 221--227.

\bibitem{shiu1987calibration}
Y.~Shiu and S.~Ahmad, ``Calibration of wrist-mounted robotic sensors by solving
  homogeneous transform equations of the form ax=xb,'' \emph{IEEE Transactions
  on Robotics and Automation}, vol.~5, no.~1, pp. 16--29, 1989.

\bibitem{shah2013solving}
M.~Shah, ``Solving the robot-world/hand-eye calibration problem using the
  kronecker product,'' \emph{Journal of Mechanisms and Robotics}, vol.~5,
  no.~3, p. 031007, 2013.

\bibitem{li2010simultaneous}
A.~Li, L.~Wang, and D.~Wu, ``Simultaneous robot-world and hand-eye calibration
  using dual-quaternions and kronecker product,'' \emph{International Journal
  of Physical Sciences}, vol.~5, no.~10, pp. 1530--1536, 2010.

\bibitem{tabb2017solving}
A.~Tabb and K.~M.~A. Yousef, ``Solving the robot-world hand-eye (s) calibration
  problem with iterative methods,'' \emph{Machine Vision and Applications},
  vol.~28, no.~5, pp. 569--590, 2017.

\bibitem{pedrosa2021general}
E.~Pedrosa, M.~Oliveira, N.~Lau, and V.~Santos, ``A general approach to
  hand--eye calibration through the optimization of atomic transformations,''
  \emph{IEEE Transactions on Robotics}, 2021.

\bibitem{collins2014infinitesimal}
T.~Collins and A.~Bartoli, ``Infinitesimal plane-based pose estimation,''
  \emph{International Journal of Computer Vision ({IJCV})}, vol. 109, no.~3,
  pp. 252--286, 2014.

\bibitem{lepetit2009epnp}
V.~Lepetit, F.~Moreno-Noguer, and P.~Fua, ``Epnp: An accurate o (n) solution to
  the pnp problem,'' \emph{International journal of computer vision}, vol.~81,
  no.~2, p. 155, 2009.

\bibitem{kneip2014upnp}
L.~Kneip, H.~Li, and Y.~Seo, ``Upnp: An optimal o (n) solution to the absolute
  pose problem with universal applicability,'' in \emph{European Conference on
  Computer Vision}.\hskip 1em plus 0.5em minus 0.4em\relax Springer, 2014, pp.
  127--142.

\bibitem{opencv_library}
G.~Bradski, ``{The OpenCV Library},'' \emph{Dr. Dobb's Journal of Software
  Tools}, 2000.

\end{thebibliography}
}

\end{document}